\documentclass{article}
\pdfoutput=1
\PassOptionsToPackage{numbers, compress}{natbib}

    \usepackage[preprint]{neurips_2020}

\usepackage[utf8]{inputenc} 
\usepackage[T1]{fontenc}    
\usepackage{hyperref}       
\usepackage{url}            
\usepackage{booktabs}       
\usepackage{amsfonts}       
\usepackage{nicefrac}       
\usepackage{microtype}      
\usepackage{amsmath,graphicx}
\usepackage{multirow}
\usepackage{tabu}
\DeclareMathOperator{\trace}{tr}
\DeclareMathOperator{\diag}{diag}
\usepackage{algorithm}
\usepackage[noend]{algpseudocode}

\usepackage[caption=false]{subfig}
\usepackage{array}
\usepackage{tabularx}
\usepackage{subfig}
\usepackage{wrapfig}
\usepackage[table]{xcolor}
\usepackage[numbers]{natbib}
\usepackage{bbm}
\usepackage{amsthm}
\newtheorem{theorem}{Theorem}
\newtheorem*{theorem*}{Theorem}

\title{A Feature-map Discriminant Perspective for\\ Pruning Deep Neural Networks}

\author{%
  Zejiang Hou, Sun-Yuan Kung\\
  Department of Electrical Engineering\\
  Princeton University\\
  \texttt{zejiangh,kung@princeton.edu} \\
}

\begin{document}

\maketitle

\begin{abstract}
Network pruning has become the de facto tool to accelerate deep neural networks for mobile and edge applications. Recently, feature-map discriminant based channel pruning has shown promising results, as it aligns well with the CNN's objective of differentiating multiple classes and offers better interpretability of the pruning decision. However, existing discriminant-based methods are challenged by computation inefficiency, as there is a lack of theoretical guidance on quantifying the feature-map discriminant power. 
In this paper,
we present a new mathematical formulation to accurately and efficiently quantify the feature-map discriminativeness, which gives rise to a novel criterion, \textit{Discriminant Information} (DI). We analyze the theoretical property of DI, specifically the non-decreasing property, that makes DI a valid selection criterion. DI-based pruning removes channels with minimum influence to DI value, as they contain little information regarding to the discriminant power. The versatility of DI criterion also enables an intra-layer mixed precision quantization to further compress the network. Moreover, we propose a DI-based greedy pruning algorithm and structure distillation technique to automatically decide the pruned structure that satisfies certain resource budget, which is a common requirement in reality. Extensive experiments demonstrate the effectiveness of our method: our pruned ResNet50 on ImageNet achieves 44\% FLOPs reduction without any Top-1 accuracy loss compared to unpruned model.
\end{abstract}

\section{Introduction}
Despite the empirical success of deep convolutional neural networks (CNN) on many computer vision tasks, huge model size and high computational complexity impede these deep networks’ deployment on mobile/embedded devices characterized by stringent resource constraints (e.g. energy, storage, latency). One popular approach to alleviate these problems is to utilize a model compression algorithm, which aims to generate a compact and efficient model by reducing redundancy from the deep model while maintaining the accuracy as much as possible.

Model compression can be roughly divided into four categories: low-rank decomposition \citep{jaderberg2014speeding}, quantization \citep{rastegari2016xnor,jacob2018quantization,wang2019haq}, efficient convolution operators \citep{han2019ghostnet} and filter/channel pruning \citep{li2016pruning,he2017channel}. 
In this work, we focus on channel pruning to remove redundant feature-maps and corresponding convolution filters from original CNNs. In addition, we also incorporate orthogonal approaches such as quantization on top of pruning to generate more compact models.

The core of channel pruning is to determine which channels to select or remove. A evaluation function is required to estimate the channel importance to the network prediction. Most existing methods resort to heuristics defined on the filter weights \citep{li2016pruning,he2019filter,chin2019legr}. As pointed out by \citep{lin2020hrank}, these heuristics ignore the distribution of the input images as well as the output labels, thus they may not accurately capture the channel importance to the prediction. In contrast, we propose to define the evaluation function on the feature-maps, since they can reflect both the filter properties and how the input images are transformed to predict the final labels. 
Recently, the notion of evaluating feature-map discriminant power has greatly fostered channel pruning performance. Prior art \citep{zhuang2018discrimination} introduced auxiliary cross-entropy losses to intermediate layers to select discriminative feature-maps. However, due to the additional auxiliary losses, the required retraining step is heavy in both computation time and human labor. This urges us to develop theoretical guidance on how to quantify the feature-map discriminant power so that we can employ discriminant-based channel pruning more effectively.

\textbf{Our Contributions.} (1) We take the first step towards adapting a variety of classical two-class discriminant metrics to multi-class by leveraging the one-vs-rest strategy and systematically study their applicability to channel pruning. Unfortunately, we observe that these generalized discriminant metrics cannot yield satisfactory pruning performance. This suggests that two-class discriminant-based metrics may not be accurate enough to quantifying the feature-maps discriminant power in multi-class CNN classification problems. (2) In order to quantify the multi-class feature-maps discriminativeness, we derive a novel criterion, \textit{Discriminant Information} (DI), from both perspectives of discriminant analysis and predictor learning. Theoretically, we prove the non-decreasing property of DI, which makes DI a valid selection criterion. Although DI is a data-dependent criterion, we find it highly stable and robust to the input samples distribution, meaning that we only need a small portion of training samples to accurately estimate the channel ranking. Without any additional auxiliary losses, constraints, or sensitivity analysis, DI-based pruning consistently outperforms previous state-of-the-art pruning methods. Based on the same criterion, we also propose a intra-layer mixed precision quantization to further compress the CNN models. (3) Moreover, we propose a \textit{DI-based greedy pruning} algorithm to prune the network to meet certain resource budget (e.g. FLOPs). Our algorithm automatically decides the target pruned architecture in a highly efficient manner and is able to capture the relative layer importance as well. To further accelerate our pruning process on very deep networks, we also propose a \textit{structure distillation} technique to transfer the patterns of pruned structures from shallower networks to compressing deeper networks on the same or different datasets. (4) Experiments on multiple datasets (CIFAR10/100, ImageNet, CUB-200) using various deep neural networks (e.g. VGG, ResNet, MobileNet) show that all the proposed strategies in our DI-based pruning method contributes to the significant performance improvement compared to other state-of-the-art methods.

\section{Related works}
Channel pruning (a.k.a. filter pruning) can be empirically categorised into four groups. \textbf{(1)} Score-based methods assign an importance score to each channel/filter of a pre-trained network and remove those below a certain threshold. After pruning, the accuracy is compensated via finetuning. Many of prior arts \citep{li2016pruning,he2018soft,yang2018netadapt,chin2019legr} take filter norm as the importance measure. \citep{he2019filter} analyzed the limitation of norm-based pruning methods and proposed a geometric median based filter pruning method. \citep{yu2018nisp} proposed a feature selection method to score the neurons in final response layer, and propagated the importance score back to preceding layers by assuming Lipschitz continuity. \citep{lin2018accelerating,you2019gate} estimated the neuron importance by applying Taylor expansion to the final cross-entropy loss. Recently, \citep{lin2020hrank} proposed to use feature-map rank as the importance measure and chose to prune filters generating low-rank feature-maps. \textbf{(2)} Training-based methods perform channel selection and training/finetuning jointly, usually by imposing extra sparsity constraints the loss function. For example, \citep{liu2017learning,wang2019pruning} employed LASSO constraint on the scaling factors in batch-normalization. \citep{huang2018data,lin2019towards,li2019compressing} introduced extra binary masks to convolutions and imposed LASSO constraints onto the masks during CNN training. Training-based pruning methods demands cumbersome hyper-parameters tuning because the modified loss function usually requires specialized optimizer such as FISTA or ADMM. \textbf{(3)} Reconstruction-based methods seek to minimize the discrepancy between the intermediate feature-maps of the pruned network and those of the original network. Many of them perform channel selection in a layer-wise manner. \citep{he2017channel,guo2020channel} conducted LASSO-based channel selection and least-square regression to minimize reconstruction error. \citep{luo2017thinet} proposed a greedy algorithm to minimize the reconstruction error. \textbf{(4)} Our method belongs to the family of feature discriminativeness based channel pruning, which is rarely explored in previous methods. The most related work is the discriminant-aware pruning (DCP) \citep{zhuang2018discrimination}. \textbf{Our work is different by}: (i) DCP relies heavily on auxiliary losses and enormous retraining to perform channel selection, while we propose an analytical criterion to efficiently quantify feature discriminativeness and select channels solely based on DI, i.e. we eliminate the need of additional losses/constraints and retraining, (ii) DCP performs channel pruning in a layer-wise fashion with repeated finetuning, while our method prunes unimportant channels across different layers simultaneously and only needs to finetune the final model by few epochs.

\section{Method}

\subsection{Discriminant-based heuristics for channel selection}
The goal of our pruning method is to maximally preserve the feature discriminant power and remove those channels with minimum contribution to the feature discriminant power. It is imperative to define a metric to efficiently quantify the feature discriminant power of intermediate layers. Many of discriminant-based criteria have been proposed, including signed SNR \citep{golub1999molecular}, Fisher discriminant ratio \citep{pavlidis2001gene}, symmetric divergence \citep{mak2006solution}, and two-sample t-test \citep{student1908probable}. However, these criteria are only applicable to measure how effectively the feature may be used to differentiate two classes. Although one may leverage one-vs-rest strategy to extend these criteria to multi-class scenarios, our empirical analysis suggest that these criteria perform worse than our proposed DI criterion (directly designed for multi-class discriminant) when they are applied to network pruning. We present detailed analysis and comparison in the supplementary.

\subsection{Quantifying feature-map discriminant power based on DI}
Quantifying discriminativeness is a classical problem, and can be approached from two perspectives: (1) discriminant analysis, which was advocated by Fisher'1936 \cite{fisher1936use} and Rao'1948 \cite{rao1948utilization}, (2) characterizing how well the feature can predict the label in the context of a specific predictor learning \cite{alain2016understanding}. We will show that \textit{Discriminant Information} (DI) metric can be interpreted from both perspectives.

\noindent\textbf{Notation.} Consider a pre-trained $L$-layer network with weights $\{\mathbf{w}^l,...,\mathbf{w}^L\}$, where $\mathbf{w}^l$ represents the filters in $l^{th}$ layer, $\mathbf{w}^l_{j,:}, j\in[C^l]$ indicates the filter for $j^{th}$ channel in $l^{th}$ layer, and $C^{l}$ is the number of filters/channels. Given an input sample $\mathcal{I}$, let $\mathbf{x}^l(\mathcal{I})$ denote the feature-maps \footnote{Feature-maps refer to activations of the layer, i.e. after convolution, batch-normalization and ReLU.} of $l^{th}$ layer, $\mathbf{x}^l_{j,:}(\mathcal{I}), j\in[C^l]$ indicates the $j^{th}$ channel which is generated by filter $\mathbf{w}^l_{j,:}$. For each layer, channel pruning aims to identify unimportant set of channels $\mathcal{T}^l$ ($|\mathcal{T}^l|=\kappa^lC^l$,$\kappa^l$ is channel sparsity) such that filters $\mathbf{w}^l_{j,:},j\in\mathcal{T}$ and feature-maps $\mathbf{x}^l_{j,:}(\mathcal{I}),j\in\mathcal{T}$ are removed. By introducing a binary indicator $\mathbf{m}^l\in\{0,1\}^{C^l}$ ($\|\mathbf{m}^l\|_0=\kappa^lC^l$) to the feature-maps (i.e. $\mathbf{x}^l(\mathcal{I})\odot\mathbf{m}^l$), pruning $j^{th}$ channel is equivalent to setting $m^l_j=0$. This also implies that the corresponding $\mathbf{w}^l_j$ filter is pruned. 

\noindent\textbf{Perspective of discriminant analysis.} General discriminant analysis aims to find a feature (sub)space that maximizes the feature signal-to-noise ratio. For ease of presentation, we will henceforth assume that the layer is fully-connected, and extension to convolution layer is presented in the supplementary. In this case, the feature-maps $\mathbf{x}^l(\mathcal{I})$ of each layer will be vectors. Given $N$ input samples, let ${\mathbf{X}}^l=[{\mathbf{x}}^l(\mathcal{I}_1),...,{\mathbf{x}}^l(\mathcal{I}_N)]$ be the matrix of feature-maps of $l^{th}$ layer, and $\mathbf{Y}=[\mathbf{y}_1,...,\mathbf{y}_N]$ be the label matrix. Inspired by Fisher'1936 \cite{fisher1936use}, we propose the following discriminant analysis to quantify the feature discriminant power in the contemporary neural network:
\begin{align}
& \underset{\mathbf{F}}{\text{maximize}}~~\trace([\mathbf{F}^T(\bar{\mathbf{K}}^l+\rho\mathbf{I})\mathbf{F}]^{-1}\mathbf{F}^T\mathbf{K}_B^l\mathbf{F})\label{raylaigh-quotient}
\end{align}
where $\mathbf{K}_B^l={\mathbf{X}}^l\mathbf{C}\mathbf{Y}^T\mathbf{Y}\mathbf{C}^T{\mathbf{X}}^{lT}$ represents the feature signal matrix, $\bar{\mathbf{K}}^l={\mathbf{X}}^l\mathbf{C}{\mathbf{X}}^{lT}$ represents the feature noise matrix, $\mathbf{C}=\mathbf{I}-{1}/{N}\mathbf{1}\mathbf{1}^T$ is the centering matrix, and $\mathbf{F}$ is the basis of the feature (sub)space w.r.t. which we optimize. $\rho$ ensures numerical stability. Problem \ref{raylaigh-quotient} belongs to the generalized Rayleigh quotient, which has a standard solution $\mathbf{F}^*=(\bar{\mathbf{K}}^l+\rho\mathbf{I})^{-1}{\mathbf{X}}^l\mathbf{C}\mathbf{Y}^T$. Upon plugging in $\mathbf{F}^{*}$ to Eq.\eqref{raylaigh-quotient}, we obtain a closed-form formula of the maximum feature signal-to-noise ratio, which we refer to as DI:
\begin{align}
& DI = \trace((\bar{\mathbf{K}}+\rho\mathbf{I})^{-1}\mathbf{K}_B)\label{DI}
\end{align}

\noindent\textbf{Perspective of predictor learning.} By learning a predictor on the intermediate layer of the network, we can characterize how well the feature predicts the label, i.e. the performance of such predictor serve as a proxy for the feature discriminativeness. We opt for those predictors with close-form solutions in order to efficiently quantify the discriminativeness without the need of gradient-based training. Accordingly, we use ridge regressor to characterize the feature discriminativeness:
\begin{align}
& \underset{\mathbf{F},\mathbf{b}}{\text{minimize}}~~\|\mathbf{F}^T{\mathbf{X}}^l+\mathbf{b}\mathbf{1}^T-\mathbf{Y}\|_F^2+\rho\|\mathbf{F}\|_F^2\label{LSE}
\end{align}
Setting the gradient of Eq.\eqref{LSE} to zero yields the optimal bias vector $\mathbf{b}^{*} = N^{-1}(\mathbf{Y}\mathbf{1}-\mathbf{F}^T{\mathbf{X}}^l\mathbf{1})$ and the optimal weight matrix $\mathbf{F}^{*} =(\bar{\mathbf{K}}^l+\rho\mathbf{I})^{-1}{\mathbf{X}}^l\mathbf{C}\mathbf{Y}^T $. Upon plugging in $\mathbf{F}^*,\mathbf{b}^*$, we obtain the minimum ridge least-square error (MRLSE):
\begin{align}
& MRLSE = -\trace((\bar{\mathbf{K}}^l+\rho\mathbf{I})^{-1}\mathbf{K}_B^l)+\|\mathbf{Y}\mathbf{C}\|_F^2=-DI+\|\mathbf{Y}\mathbf{C}\|_F^2
\end{align}
Ignoring the constant term $\|\mathbf{Y}\mathbf{C}\|_F^2$, MRLSE can be minimized by maximizing the DI value.

Therefore, DI has two mathematical interpretations. Higher DI value means that (1) the feature contains little noise but discriminant information for prediction; (2) the feature can better predict the label with smaller MRLSE. Before we apply DI to channel pruning, we discuss the theoretical property that makes DI a valid selection criterion, where the proof is presented in the supplementary:
\begin{theorem}[Non-decreasing of DI]
Denote $\mathcal{T},\mathcal{S}$ as the set of features corresponding to some channels in $l^{th}$ layer of the network. We have $DI(\mathcal{T})\leq DI(\mathcal{S})$ whenever $\mathcal{T}\subseteq\mathcal{S}$. DI achieves its maximum for the full channel set, i.e. no pruning.\label{non-decreasing}
\end{theorem}

\subsection{DI-based Channel Pruning}
The next question would be applying DI to channel pruning. Given a desired channel sparsity $\kappa^{l}$, we need to find $\kappa^lC^l$ channels that maximally preserve the feature discriminant power, i.e. DI value. A straight-forward way is perform exhaustive search, but it requires to evaluate $\binom{C^l}{\kappa^lC^l}$ times to do just one layer of pruning, which is computationally prohibitive for current deep and wide networks.

Due to the binary indicator vectors $\mathbf{m}^{l}$, we have separated the activation of the channel from whether the channel is present or not. Therefore, we may be able to determine the channel importance by measuring its influence on the DI value, i.e. to measure the difference in DI value when $m^{l}_j=1$ (channel $j$ is present) and $m^{l}_j=0$ (channel $j$ is pruned) while keeping everything else constant. Precisely, the influence of removing $j^{th}$ channel on the DI value can be measured by:
\begin{align}
& \phi^{l}_j = \trace\{[\diag(\mathbf{1})\bar{\mathbf{K}}^l\diag(\mathbf{1})+\rho\mathbf{I}]^{-1}\diag(\mathbf{1})\mathbf{K}_{B}^l\diag(\mathbf{1})\}-\\\nonumber
& \trace\{[\diag(\mathbf{1}-\mathbf{e}_j)\bar{\mathbf{K}}^l\diag(\mathbf{1}-\mathbf{e}_j)+\rho\mathbf{I}]^{-1}\diag(\mathbf{1}-\mathbf{e}_j)\mathbf{K}_{B}^l\diag(\mathbf{1}-\mathbf{e}_j)\}
\end{align}
where $\mathbf{e}_j$ has zeros everywhere except for $j^{th}$ element where it is one, and $\diag(.)$ converts a vector to diagonal matrix. By relaxing the binary constraint on the indicator vector $\mathbf{m}^{l}$, $\phi^{l}_j$ can be approximated by the derivative of DI with respect to $m^l_j$. The derivative can be succinctly expressed in the following formula, where the proof is presented in supplementary:
\begin{align}
& \phi^{l}_j\approx\dfrac{\partial \trace\{[\diag(\mathbf{m}^l)\bar{\mathbf{K}}^l\diag(\mathbf{m}^l)+\rho\mathbf{I}]^{-1}\diag(\mathbf{m}^l)\mathbf{K}_{B}^l\diag(\mathbf{m}^l)\}}{\partial m^{l}_j}\bigg|_{\mathbf{m}^l=\mathbf{1}}=2\rho([\bar{\mathbf{K}}^l]^{-1}\mathbf{K}_B^l[\bar{\mathbf{K}}^l]^{-1})_{jj}
\end{align}
$\partial DI/\partial m^l_j$ is an infinitesimal version of $\phi^l_j$, that measures the change of DI value due to a multiplicative perturbation to the activation of the channel. When the value of the derivative is high, $j^{th}$ channel has a considerable influence on the feature discriminant power in $l^{th}$ layer, so it has to be preserved. After we compute $\phi^l_j$ for all channels and sort them in a descending order, i.e. $Rank^l=sort(\phi^l_1,...,\phi^l_{C^l})$, channel pruning is performed by setting $m^l_j=\mathbbm{1}\{j\in Rank^l[:\kappa^lC^l]\}, j\in[C^l]$, i.e. retaining filters generating top-ranked feature-maps while removing those generating bottom-ranked feature-maps. We evaluate and prune unimportant filters across different layers simultaneously.

\begin{figure}[t]
\captionsetup[subfigure]{labelformat=empty, font=scriptsize}
\centering
\small
\includegraphics[scale=0.2]{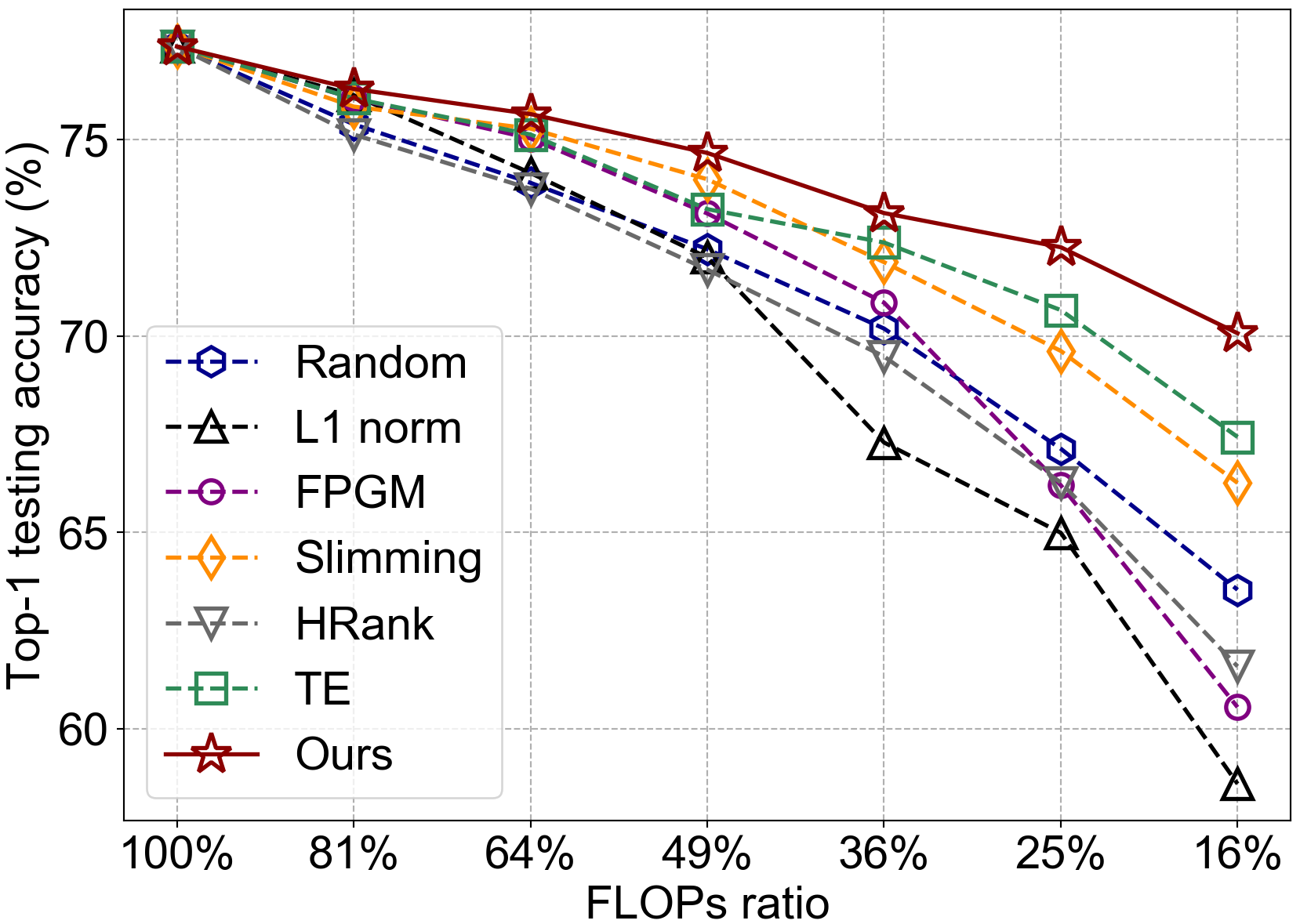}\hspace{0.4in}
\includegraphics[scale=0.35]{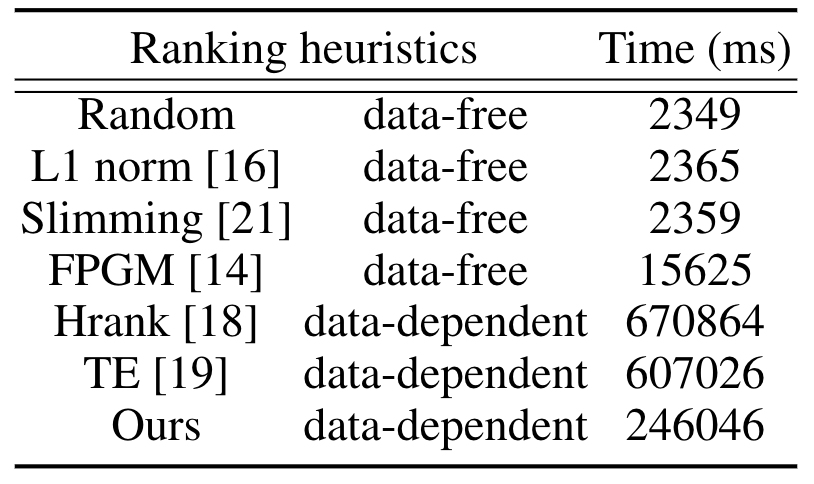}
\vspace{-0.1in}
\caption{Comparison of channel selection heuristics in terms of testing accuracy (left figure) and computation time (right figure). Computation cost $=$ scoring all channels of whole network $+$ sorting $+$ thresholding $+$ pruning. We prune VGG16 on CUB-200 dataset with different FLOPs reduction ratios. Results are averaged over three independent replicates.}
\label{figs:ranking heuristic comparison}
\vspace{-0.15in}
\end{figure}

\noindent\textbf{Comparison of channel selection heuristics.}
We compare DI with five recently proposed ranking heuristics, including filter norm \citep{li2016pruning}, geometric median (FPGM) \citep{he2019filter}, Taylor expansion (TE) \citep{lin2018accelerating}, batch-normalization scaling factor (Slimming) \citep{liu2017learning}, and random selection. We adopt VGG16 \citep{simonyan2014very} as the baseline network and evaluate on the widely used fine-grained CUB-200 \citep{wah2011caltech} dataset. All pruning tests are conducted on the same baseline. For the purpose of fair comparison, we simply prune each layer by the same ratio (e.g. 10\%, 20\%,...,60\%) according to different ranking heuristics, and all pruned models are retrained by only 10 epochs with learning rate $1\times10^{-3}$. From Fig.\ref{figs:ranking heuristic comparison}, DI-based pruning obtains better accuracy across different FLOPs reduction ratios, and the improvement becomes more significant when the network is aggressively pruned. Moreover, Fig.\ref{figs:ranking heuristic comparison} shows the computation time of different methods for channel selection. We observe that data-free heuristics ($\ell_1$-norm, FPGM, Slimming, Random) are more efficient but at cost of worse accuracy. Our method is significantly faster than TE because computing DI only requires forward propagation while TE requires both forward and backward propagation.


\subsection{Pruning to meet resource budget}
In reality, we are interested in pruning the network to meet certain resource budget (e.g. FLOPs) while maximizing the accuracy. Previous methods \citep{he2018soft,he2019filter} tend to prune every layer with the same ratio, and obtain pruned architectures satisfying different resource budgets by varying the ratio. However, the assumption that each layer is of equally important to accuracy is hardly true in contemporary CNNs \citep{liebenwein2019provable}. To address it, we propose a greedy pruning algorithm to meet the resource goal progressively, and the algorithm pseudo-code is provided in Algorithm 1. At each step, two operations are applied to each layer (convolution or fully-connected) individually: (i) determine how many channels and filters to remove in the layer so that the pruned model can reduce the resource by $\delta$ (referred to as reduction schedule, e.g. $\delta$=0.5\% FLOPs);
(ii) obtain a subnet by pruning bottom-ranked channels and filters based on the DI criterion.
As a result, each step will generate $L$ subsnets, each of which has only one layer pruned and reduces the resource by $\delta$. We evaluate these subnets on a validation set\footnote{We split 10\% of the original training set and use it as the validation set.}. The one with the highest validation accuracy is carried over to next step. The process ends when the network resource consumption $Res^{t}$ meets the resource goal $\phi$, after which we finetune the final model to recover the accuracy, i.e. we eliminate the need of iterative pruning-finetuning cycles. 
In the supplementary, we illustrate how our greedy pruning automatically decides the target pruned architecture
and show that our algorithm can capture the relative layer importance as well. 

\begin{figure}[t]
\captionsetup[subfigure]{labelformat=empty, font=scriptsize}
\centering
\small
\subfloat[]{\includegraphics[scale=0.2]{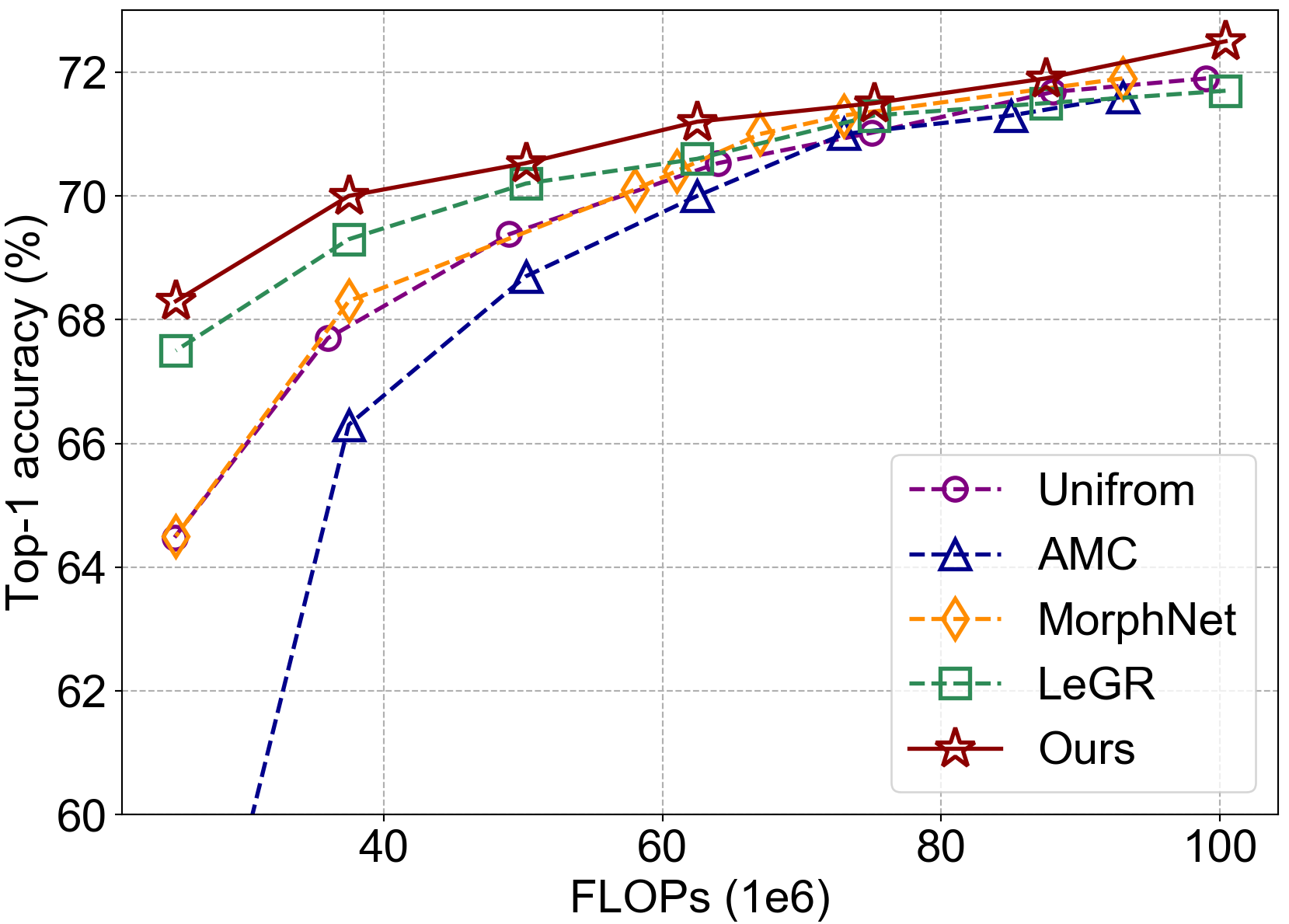}}\hspace{0.4in}
\subfloat[]{\includegraphics[scale=0.2]{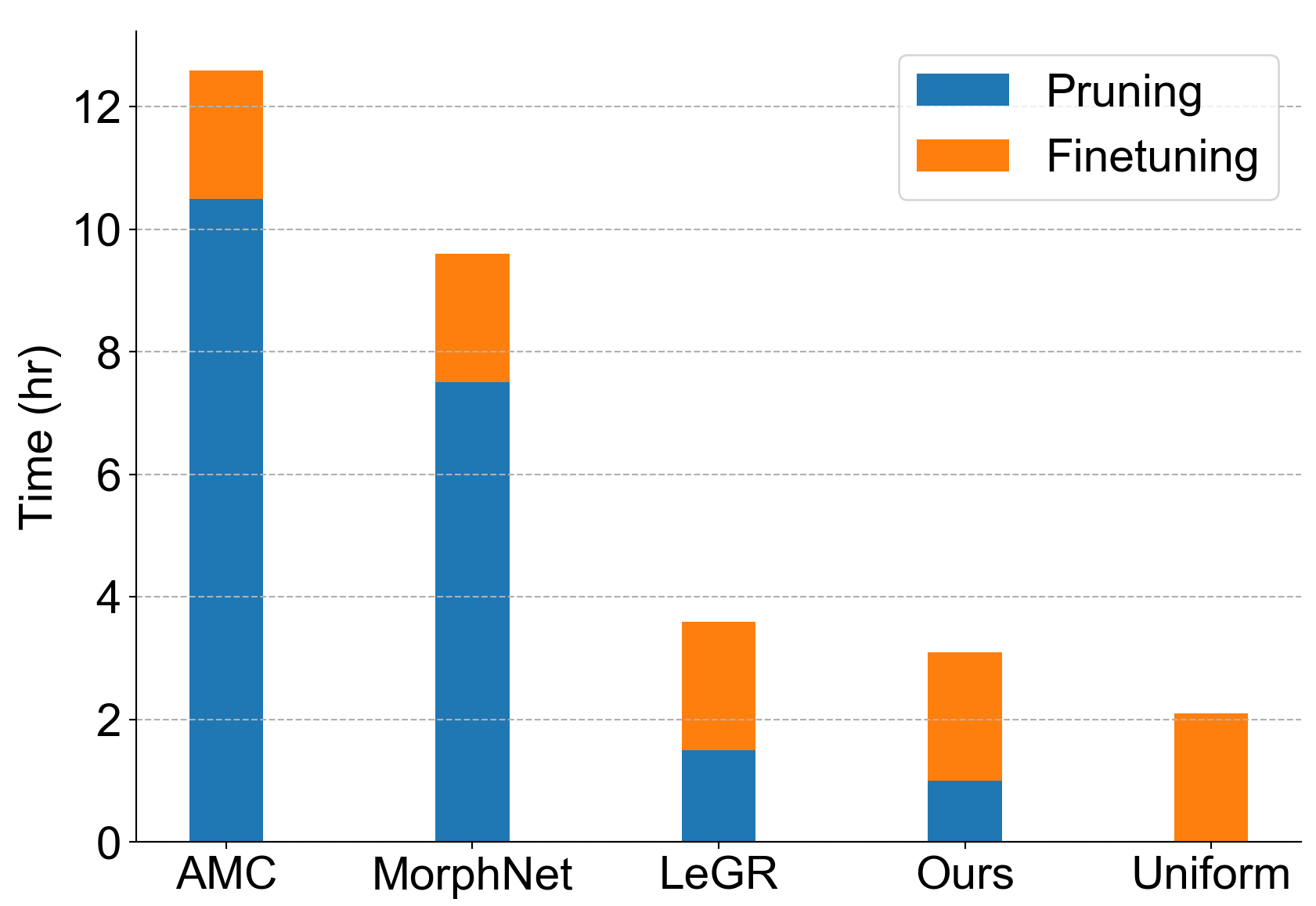}}
\vspace{-0.2in}
\caption{Left: accuracy vs. FLOPs curve of pruning ResNet56 on CIFAR100 dataset. Right: computation time to obtain seven pruned ResNet56 under different FLOPs budgets using different methods. Results are averaged over three independent replicates.}
\label{figs:pruning under resource constraint}
\vspace{-0.15in}
\end{figure}

\noindent\textbf{Comparison of pruning strategies.} We consider pruning ResNet56 on CIFAR100 and compare with four leading approaches for pruning to meet resource budget, including LeGR \citep{chin2019legr}, AMC \citep{he2018amc}, MorphNet \citep{gordon2018morphnet}, and a uniform pruning baseline. 
We apply each algorithm to obtain seven pruned ResNet56 with different FLOPs reduction goals (20\%,30\%,...,80\%) and finetune them with the same procedure. In Fig.\ref{figs:pruning under resource constraint}, we evaluate in terms of final accuracy and computation time to obtain the seven pruned models. The computation time includes (1) pruning: the time it takes to obtain pruned networks satisfying the FLOPs budgets; (2) finetuning: the time it takes to finetune the pruned networks. Our pruning algorithm obtains higher accuracy than other approaches under different FLOPs budgets. More importantly, our pruning pipeline can explore the trade-off between accuracy and resource more efficiently. To explain, AMC \citep{he2018amc} employs reinforcement learning to search the pruned architectures, LeGR \citep{chin2019legr} uses evolutionary algorithm to obtain the pruned architectures, while we alleviate these searching/training overhead by directly pruning the network in a greedy fashion.

\noindent\textbf{Structure Distillation.} Although our pruning algorithm is already more efficient than the comparative methods, its computation cost would increase as the network goes deeper. On the other hand, we observe that the pruned structures are usually generalizable across architectures and datasets. Thus, we propose \textit{Structure Distillation} (SD) to further accelerate our pruning process on very deep networks. Specifically, we distill the patterns of pruned structures from shallower networks (e.g. ResNet18 on ImageNet) and apply them to compress deeper networks (e.g. ResNet50 on ImageNet) on the same or different datasets. The distillation is performed by averaging the pruning ratios in each stage of the pruned shallower networks.

\subsection{Variants of our pruning method}
In this paper, we investigate three variants of our DI-based channel pruning. \textbf{DI-unif}: we use DI to rank the channel importance and simply prune all convolution layers with the same pruning ratio (i.e. uniform pruning). Given that many existing works adopt the uniform pruning and focus on ranking heuristics comparison, we believe DI-unif can validate the effectiveness of our proposed DI ranking heuristics. \textbf{DI-greedy}: we use DI to rank the  channel importance and employ greedy algorithm to prune the network to meet certain resource budget. \textbf{DI-SD}: we use DI to rank the channel importance and employ \textit{structure distillation} to construct the pruned structures.

\subsection{Intra-layer mixed precision quantization}
The model size of pruned network can be further compressed through weight quantization. Conventional quantization methods treat filters within each layer equally and assign the same number of bits to all filters, but as different filters have different importance for prediction, it is imperative to use mixed precision for different filters of each layer. In filter pruning, we discard unimportant filters (i.e. 0-bit) while retaining important ones (i.e. 32-bit) according to DI. Similarly, in quantization, we assign less bits to unimportant filters while using more bits for important ones based on the same ranking heuristic, i.e. bi-level mixed precision. After pruning and finetuning a given deep network, we apply linear quantization with mixed precision to each convolution layer and fully-connected layer, and the quantized network is finetuned by the same procedure as \citep{wang2019haq}. Our proposed intra-layer mixed precision quantization boosts the compression ratio with negligible accuracy loss (cf. Table \ref{tab:mixed precision quantization}).

\section{Experiment}
We validate the effectiveness of our method on four benchmarks, i.e., CIFAR10/100 \citep{krizhevsky2014cifar}, ImageNet \citep{imagenet_cvpr09}, and CUB-200 \citep{wah2011caltech}, using three popular architectures, i.e. VGGNet \citep{simonyan2014very}, ResNet \citep{he2016deep}, and MobileNetV2 \citep{sandler2018mobilenetv2}. We prune every convolution layer in the network. For ResNet, we group channels that are summed by shortcut connections by summing up their importance score and prune them jointly. For MobileNetV2, the second layer in each bottleneck block contains depth-wise convolution, meaning that the input dimension should be the same as the output. Hence, the first two layers in each bottleneck block are pruned jointly. We follow the existing methods to evaluate the compressed models in terms of Top-1 testing accuracy, FLOPs, and parameters. Implementation details are summarized in the supplementary.

\subsection{Ablation studies}
We investigate different aspects of our pruning method through ablation studies. Results are based on pruning MobileNetV2 on CIFAR100. Similar observations hold for other architectures and datasets.

\begin{figure}[t]
\centering
\subfloat[]{\includegraphics[scale=0.15]{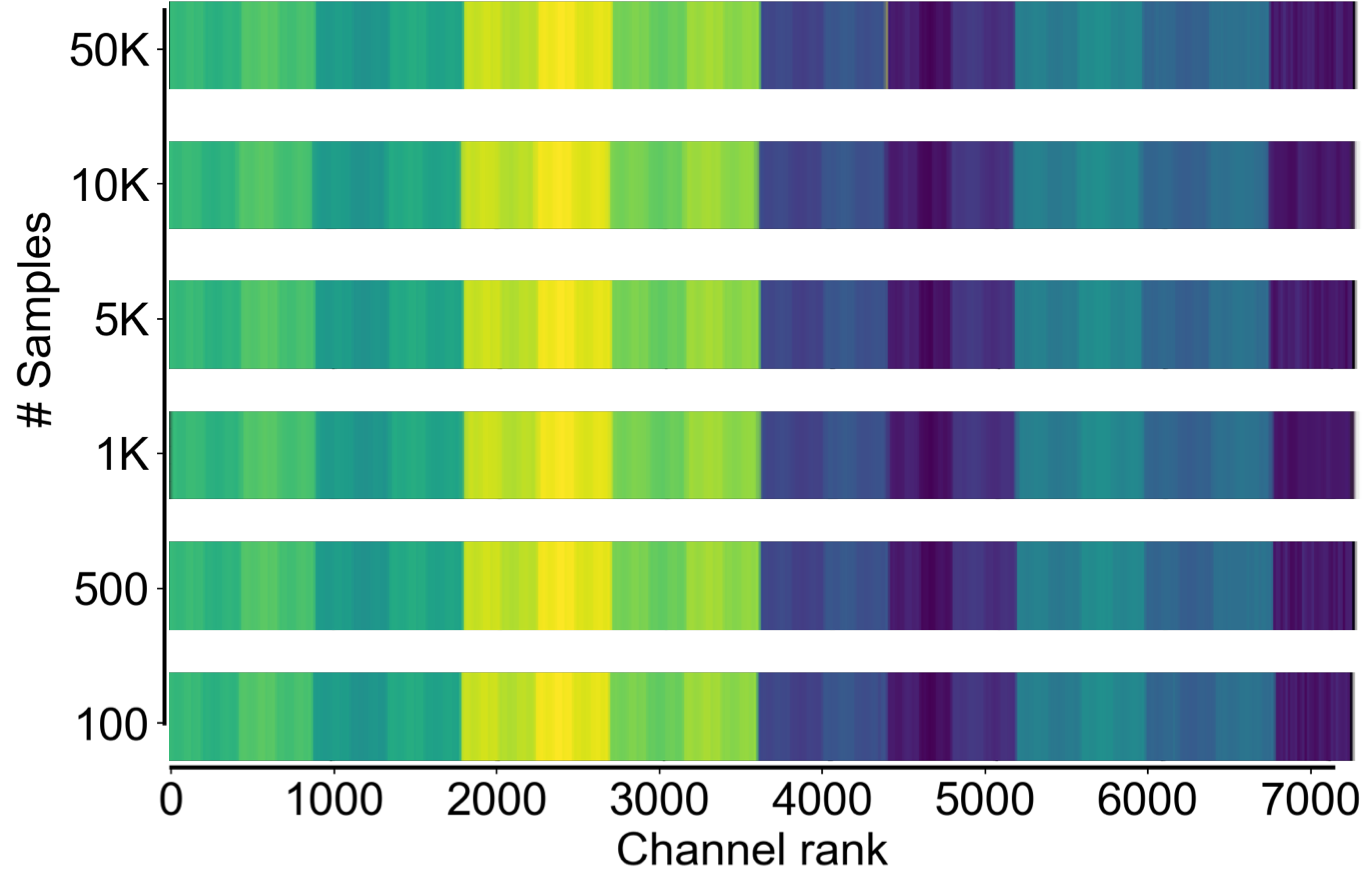}}\hspace{0.02in}
\subfloat[]{\includegraphics[scale=0.145]{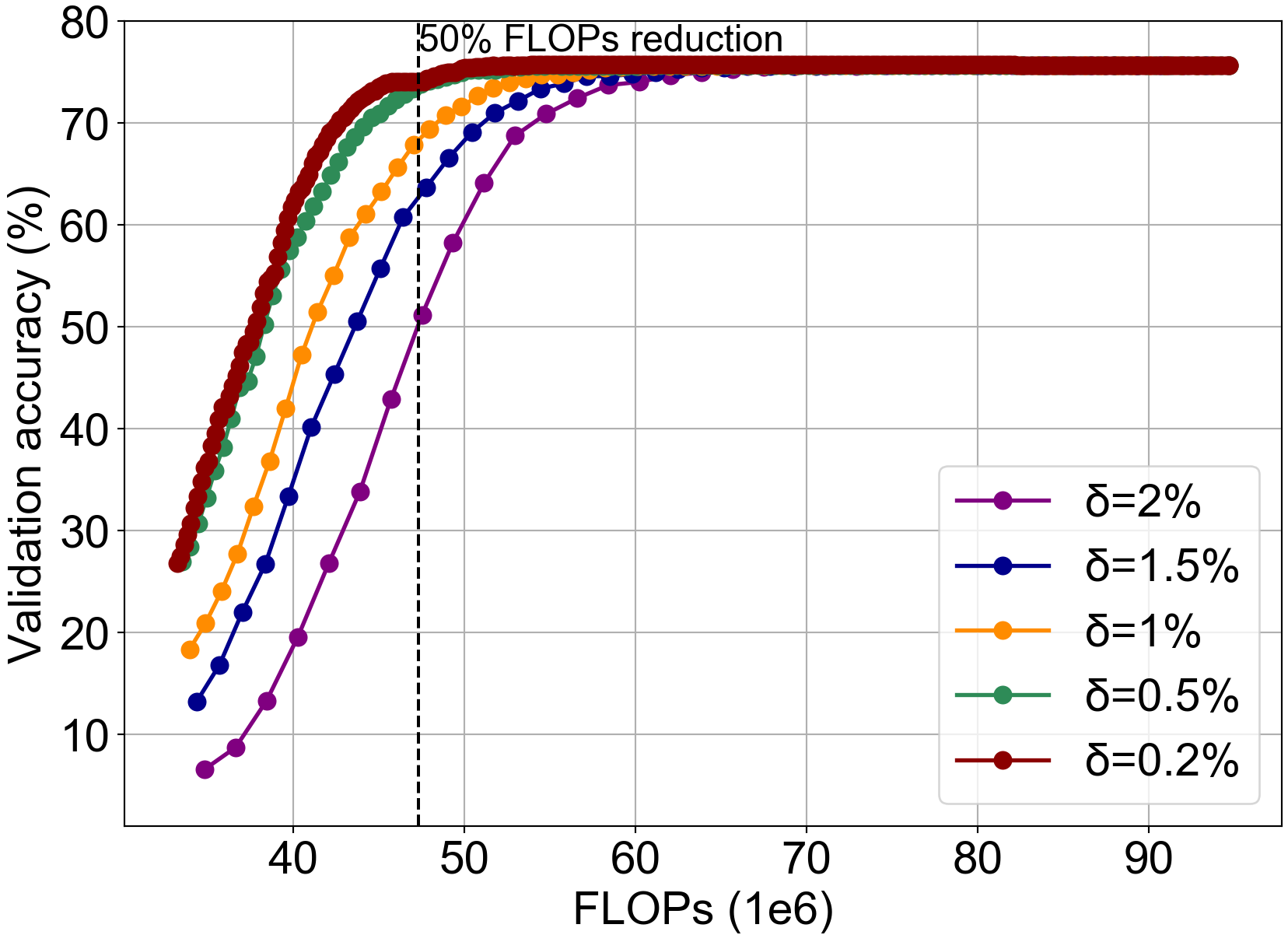}}\hspace{0.02in}
\subfloat[]{\includegraphics[scale=0.15]{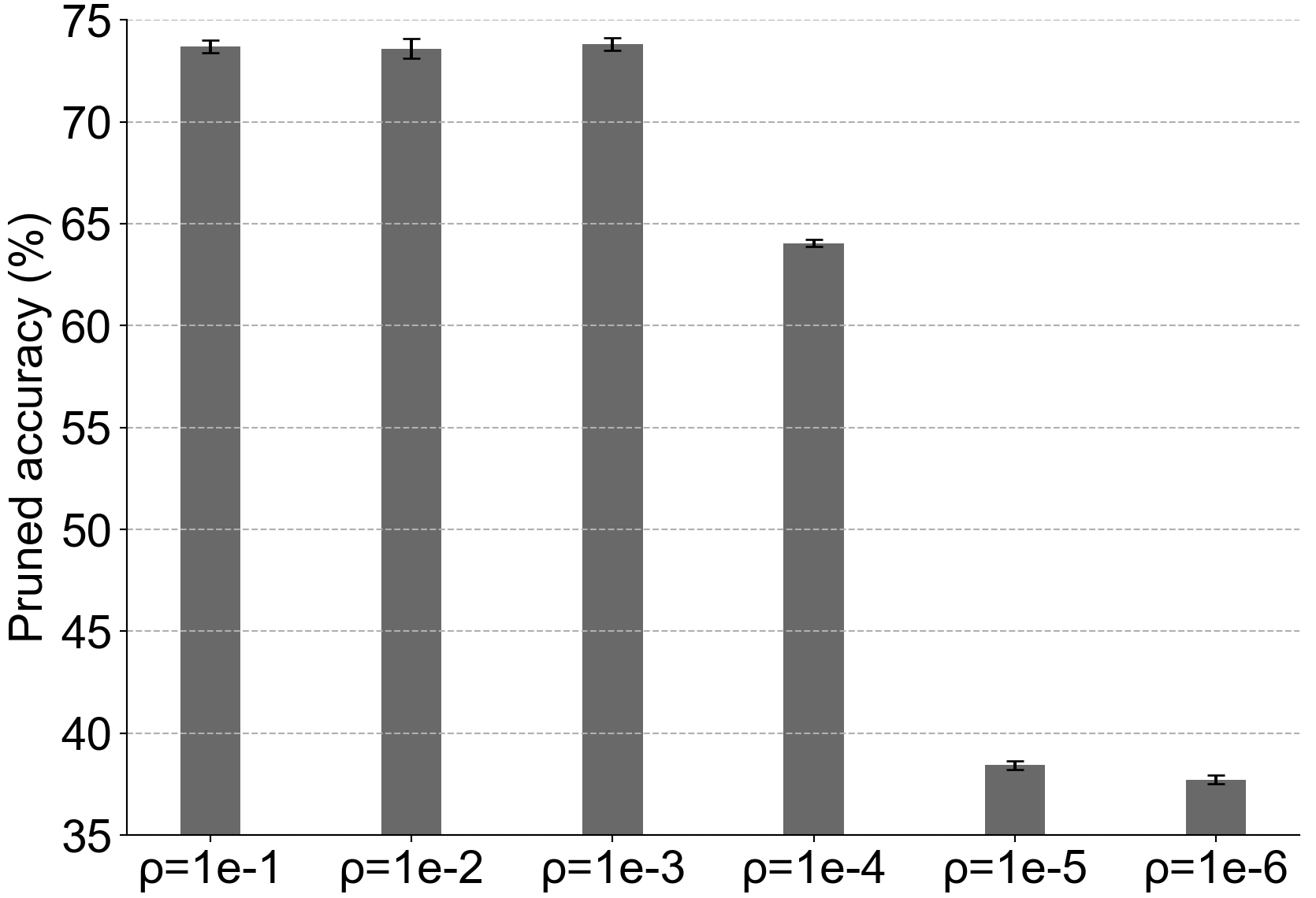}}\hspace{0.02in}
\subfloat[]{\includegraphics[scale=0.15]{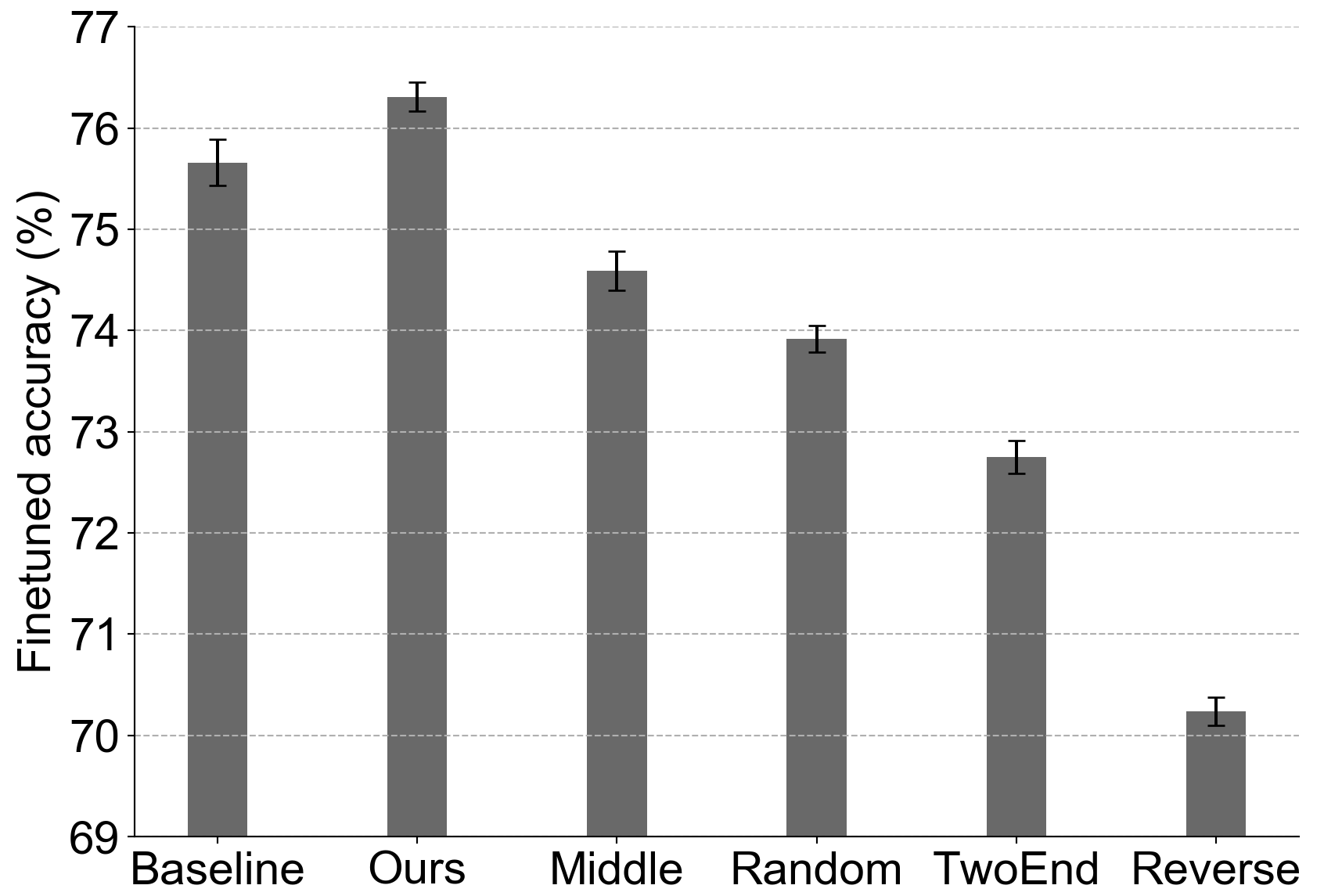}}
\vspace{-0.1in}
\caption{Ablation studies of pruning MobileNetV2 on CIFAR100. (a) Influence of number of samples on DI-based channel ranking. (b) Impact of resource reduction schedule $\delta$ on pruned accuracy on validation set. (c) Effect of regularization $\rho$ of DI on pruned accuracy on test set. (d) Comparison of different variants of DI-based selection in terms of test accuracy.}
\label{fig:ablation}
\vspace{-0.15in}
\end{figure}

\noindent\textbf{Influence of input samples on DI.} DI is a data-dependent criterion: we need to feed input samples, extract intermediate feature-maps from each layer for DI computation. Empirically, we observe that ranking channel importance based on DI is highly stable and robust to the input samples distribution. In Fig.\ref{fig:ablation}(a), we use different colors to represent the ranking of all channels globally in MobileNetV2 based on DI using different number of input samples. The ranking is almost unchanged regardless of the number of input samples used to compute DI. Thus, we can accurately and efficiently estimate the channel ranking using only a small portion of training samples.\\
\noindent\textbf{Impact of resource reduction schedule $\delta$.} In Fig.\ref{fig:ablation}(b), we investigate the impact of reduction schedule $\delta$ (as defined in Sec.3.4) on the pruned model accuracy. As seen, a smaller reduction schedule yields better pruned accuracy (i.e. accuracy of pruned model without finetuning) but requires more iterations (more dots in the plot) to meet the resource budget, e.g. 50\% FLOPs.\\
\noindent\textbf{Effect of regularization $\rho$ of DI.} $\rho$ is introduced into DI for numerical stability. We study how this parameter affects the performance of our pruning method by varying $\rho$ in range $\{10^{-1},10^{-2},...,10^{-5}\}$. The results in Fig.\ref{fig:ablation} suggests that a larger value of $\rho$ would yield better pruned accuracy. In practice, we choose $\rho=10^{-1}$ throughout our experiments.\\
\noindent\textbf{Comparison of different variants of DI-based selection.} To verify the appropriateness of retaining channels with high DI value (denoted as ``Ours"), we include four variants into comparison, including (1) Reverse: channels with high DI are pruned, (2) ``TwoEnd": channels with both high and low DI are pruned, (2) ``Middle": channels with both high and low DI are retained, (3) ``Random": channels are randomly pruned. We set the pruning rate to be the same across different variants, and report the test accuracy after pruning 50\% FLOPs2. Among four variants, ``Middle" yields the best accuracy while ``Reverse" performs the worst. This suggests that channels with low DI score indeed contain less information; but as long as channels with highest DI scores are retained, the critical information for prediction is well preserved. Moreover, we can see that ``Ours" outperforms all the other variants.\\
\noindent\textbf{DI on pre-activated vs. activated feature-maps.} Throughout our experiments, DI is computed from non-linearly activated feature-maps (``act DI"). Here, we compare with DI computed from pre-activated feature maps (``pre-act DI") or from batch-normalized feature-maps (``bn DI"). After pruning 50\% FLOPs from MobileNetV2, the test accuracy of ``act DI", ``pre-act DI", ``bn DI" are 76.31\%, 75.98\%, 75.84\%, respectively. Using ``act DI" brings in slightly better result.

\subsection{Comparison with the state-of-the-arts}
\noindent\textbf{CIFAR10/100 results} in Table \ref{tab:cifar}. 
We prune ResNet56/110/164 and VVGNet on both CIFAR10 and CIFAR100. First, we observe our most naive variant, DI-unif, yields significantly higher accuracy and higher FLOPs/parameters reduction than existing state-of-the-art methods. This shows that DI indeed selects more important channels to prediction.
Moreover, as discussed in Fig.\ref{figs:ranking heuristic comparison}, our channel pruning cost takes on the order of minutes. Thus, our pruning process is highly efficient compared with such methods as TAS (the searching cost for pruned network is 3.8 hours on CIFAR) and DCP (the channel pruning cost is 7.6 hours on CIFAR). On the other hand, DI-greedy improves the accuracy upon DI-unif and other methods while achieving higher FLOPs/parameters reduction. To explain, DI-greedy automatically decides the pruned architecture instead of using manually pre-defined rules, and it can capture the relative layer importance. Thus, we can achieve better accuracy after reducing the targeted amount of FLOPs.

\noindent\textbf{ImageNet results} in Table \ref{tab:imagenet}. 
We prune ResNet18/50 on ImageNet. We also prune more compact model, i.e. MobileNetV2 on ImageNet and result is provided in the supplementary. Again, DI-unif compares favourably against other state-of-the-art methods in terms of higher accuracy and higher FLOPs/parameter reduction. DI-unif  does not require any additional constraints, auxiliary loss, sensitivity analysis, or searching cost. Given that we finetune the pruned network with the same procedure as comparative methods, the performance improvement proves the effectiveness of DI metric on large-scale dataset. Moreover, after we distill the pruned structure of ResNet18 DI-greedy via structure distillation and apply it to prune ResNet50 (DI-SD), we manage to reduce the FLOPs by 44\% while maintaining the Top-1 accuracy as 76.10\%, whereas all other methods show significant accuracy loss. This shows that pruned patterns are indeed generalizable across different architectures and our DI-based greedy pruning algorithm can find these patterns effectively.

\begingroup
\setlength{\tabcolsep}{3pt} 
\begin{table}[t]
\small
\caption{Comparison of pruning methods on CIFAR10/100. ``Acc.": accuracy. ``FLOPs": FLOPs(pruning ratio). ``Params.": parameters(pruning ratio). ``M": million. ``-": results not reported by corresponding method.}
\vspace{-0.1in}
\centering
\begin{tabular}{c c c c c c c c}\toprule
      & \multirow{2}{*}{Method} & \multicolumn{3}{c}{CIFAR10} & \multicolumn{3}{c}{CIFAR100}\\
     & & Acc. (\%) & FLOPs (\%) & Params. (\%) & Acc. (\%) & FLOPs (\%) & Params. (\%)\\\hline\hline
     \multirow{8}{*}{R56} & SFP \citep{he2018soft} & 93.6$\rightarrow$93.4 & 59.4M(52.6) & - & 71.4$\rightarrow$68.8 & 59.4M(52.6) & -\\
     & AMC \citep{he2018amc} & 92.9$\rightarrow$91.9 & 62.5M(50.0) & - & - & - & -\\
     & DCP \citep{zhuang2018discrimination} & 93.8$\rightarrow$93.5 & 62.5M(50.0) & 0.43M(50.0) & - & - & -\\ 
     & FPGM \citep{he2019filter} & 93.6$\rightarrow$93.5 & 59.4M(52.6) & - & 71.4$\rightarrow$69.7 & 59.4M(52.6) & -\\
     & TAS \citep{dong2019network} & 94.5$\rightarrow$93.7 & 59.5M(52.7) & - & 73.2$\rightarrow$72.3 & 61.2M(51.3) & -\\
     & HRank \citep{lin2020hrank} & 93.3$\rightarrow$93.5 & 88.7M(29.3) & 0.71M(16.8) & - & - & -\\
     & \textbf{DI-unif}(Ours) & 93.6$\rightarrow$93.9 & 57.9M(53.7) & 0.41M(51.8) & 72.7$\rightarrow$72.3 & 58.0M(53.6) & 0.42M(50.6) \\
     & \textbf{DI-greedy}(Ours) & 93.6$\rightarrow$94.2 & 49.7M(60.2) & 0.43M(50.0) & 72.7$\rightarrow$72.4 & 49.7M(60.2) & 0.52M(38.8) \\\hline
     \multirow{6}{*}{R110} & SFP \citep{he2018soft} & 93.7$\rightarrow$93.9 & 150M(40.8) & - & 74.1$\rightarrow$71.3 & 121M(52.3) & -\\
     & FPGM \citep{he2019filter} & 93.7$\rightarrow$93.9 & 121M(52.3) & - & 74.1$\rightarrow$72.6 & 121M(52.3) & -\\
     & TAS \citep{dong2019network} & 95.0$\rightarrow$94.3 & 119M(53.0) & - & 75.1$\rightarrow$73.2 & 120M(52.6) & -\\
     & HRank \citep{lin2020hrank} & 93.5$\rightarrow$94.3 & 148M(41.2) & 1.04M(39.4) & - & - & -\\
     & \textbf{DI-unif}(Ours) & 93.7$\rightarrow$94.6 & 112M(55.7) & 0.75M(56.4) & 74.1$\rightarrow$73.7 & 117M(53.7) & 0.80M(53.4) \\
     & \textbf{DI-greedy}(Ours) & 93.7$\rightarrow$95.3 & 100M(60.5) & 0.83M(51.7) & 74.1$\rightarrow$74.3 & 101M(60.1) & 1.09M(36.6) \\\hline
     \multirow{4}{*}{R164} & Slimming \citep{liu2017learning} & 94.6$\rightarrow$94.9 & 190M(23.7) & 1.44M(14.9) & 76.6$\rightarrow$77.1 & 166M(33.3) & 1.46M(15.5) \\
     & TAS \citep{dong2019network} & 95.5$\rightarrow$94.0 & 178M(28.1) & - & 78.3$\rightarrow$77.8 & 171M(30.9) & - \\
     & \textbf{DI-unif}(Ours) & 94.6$\rightarrow$95.1 & 109M(56.4) & 0.71M(58.2) & 76.9$\rightarrow$77.8 & 109M(56.4) & 0.74M(57.2)\\
     & \textbf{DI-SD}(Ours) & 94.6$\rightarrow$95.7 & 98M(60.6) & 0.90M(47.1) & 76.9$\rightarrow$78.0 & 161M(35.6) & 1.27M(26.6)\\\hline
    \multirow{5}{*}{VGG} & Slimming \citep{liu2017learning} & 93.7$\rightarrow$93.8 & 195M(51.0) & 2.30M(88.5) & 73.3$\rightarrow$73.5 & 250M(37.1) & 5.00M(75.1) \\
     & DCP \citep{zhuang2018discrimination} & 94.0$\rightarrow$94.5 & 139M(65.0) & 1.28M(93.6) & - & - & - \\
     & AOFP \citep{ding2019approximated} & 93.4$\rightarrow$93.8 & 215M(46.1) & - & - & - & - \\
     & HRank \citep{lin2020hrank} & 93.9$\rightarrow$93.4 & 145M(63.6) & 2.51M(87.5) & - & - & -\\
     & \textbf{DI-greedy}(Ours) & 93.9$\rightarrow$94.7 & 91M(77.2) & 1.21M(94.0) & 73.6$\rightarrow$74.1 & 159M(60.1) & 4.13M(79.4) \\
     \bottomrule
\end{tabular}
\label{tab:cifar}
\vspace{-0.2in}
\end{table}
\endgroup

\subsection{Quantization results}
\begin{wraptable}{r}{6.5cm}
\vspace{-0.25in}
\begingroup
\setlength{\tabcolsep}{1pt} 
\scriptsize
\caption{Comparison of quantization results for ResNet50 on ImageNet.}
\centering
\begin{tabular}{c c c c c}\toprule
     Method & Precision & Model Size & Top-1 & Top-5 \\\hline\hline
    ResNet50 & float 32-bit & 97.49 MB & 76.10\% & 92.93\% \\
    DC \citep{han2015deep} & fixed 2-bit & 6.32 MB & 69.85\% & 88.68\% \\
    FSNet-WQ \citep{yang2019fsnet} & fixed 8-bit & $\approx$8.37 MB & 69.87\% & 89.61\% \\
    HAQ \citep{wang2019haq} & inter-layer mixed & 6.30 MB & 70.63\% & 89.93\% \\
    Ours & intra-layer mixed & 6.09 MB & \textbf{73.21\%} & \textbf{91.27\%} \\
    \bottomrule
\end{tabular}
\label{tab:mixed precision quantization}
\endgroup
\vspace{-0.15in}
\end{wraptable}
In Table \ref{tab:mixed precision quantization}, we compare our intra-layer mixed precision quantization with Deep Compression (DC) \citep{han2015deep}, Hardware-Aware Automated Quantization (HAQ) \citep{wang2019haq}, FSNet-WQ \citep{yang2019fsnet} by compressing ResNet50 \citep{he2016deep} on ImageNet \citep{imagenet_cvpr09}. Our method obtains significantly higher accuracy with similar or smaller model size. Moreover, our compressed ResNet50 is smaller and more accurate than the expert-designed compact networks, e.g. MobileNetV2 (71.8\%, 14.2 MB) and ShuffleNetV2 (69.36\%, 9.2 MB). Of note is that HAQ \citep{wang2019haq} proposes inter-layer mixed precision quantization by leveraging reinforcement learning, which incurs a great amount of computation overhead. In contrast, we propose intra-layer mixed precision quantization based on an analytical heuristic, which can be efficiently computed. 

\begingroup
\setlength{\tabcolsep}{6pt}
\begin{table}[t]
\footnotesize
\centering
\caption{Comparison of pruning results on ImageNet. ``$\Delta$Top-1": Top-1 accuracy difference due to pruning, calculated as \textit{pruned top-1 $-$ pre-trained top-1} of each method. ``FLOPs": FLOPs(pruning ratio). ``Params.": parameters(pruning ratio). ``-": results not reported by corresponding method.}
\vspace{-0.1in}
\begin{tabular}{c c c c c c}\toprule
    Model & Method & Top-1 (\%) & $\Delta$Top-1 (\%) & FLOPs & Params. \\\hline\hline
    \multirow{6}{*}{ResNet18} & SFP \citep{he2018soft} & 70.28 $\rightarrow$ 67.10 & -3.18  & 1.05E9 (41.8\%) & - \\
    & FPGM \citep{he2019filter} & 70.28 $\rightarrow$ 68.41 & -1.87 & 1.05E9 (41.8\%) & - \\
    & \textbf{DI-greedy}(Ours) & 69.76 $\rightarrow$ 68.91 & -0.85 & 1.04E9 (42.5\%) & 7.82E6 (33.1\%) \\
    & DCP \citep{zhuang2018discrimination} & 69.76 $\rightarrow$ 67.35 & -2.41 & 0.98E9 (46.1\%) & 6.19E6 (47.1\%) \\
    & Sampling \citep{liebenwein2019provable} & 69.76 $\rightarrow$ 67.38 & -2.38 & 1.28E9 (29.3\%) & 6.57E6 (43.8\%) \\
    & \textbf{DI-unif}(Ours) & 69.76 $\rightarrow$ 68.15 & -1.61 & 0.98E9 (46.1\%) & 6.19E9 (47.1\%) \\\hline
    \multirow{15}{*}{ResNet50} & SSS-32 \citep{huang2018data} &  76.10 $\rightarrow$ 74.18 & -1.92 & 2.82E9 (31.1\%) & 18.60E9 (27.3\%) \\
    & SFP \citep{he2018soft} & 76.15 $\rightarrow$ 74.61 & -1.54 & 2.38E9 (41.8\%) & - \\
    & FPGM-30 \citep{he2019filter} & 76.15 $\rightarrow$ 75.59 & -0.56 & 2.36E9 (42.2\%) & - \\
    & GAL-0.5 \citep{lin2019towards} & 76.15 $\rightarrow$ 71.95 & -4.20 & 2.33E9 (43.1\%) & 21.20E6(17.2\%) \\
    & Taylor-81 \citep{molchanov2019importance} & 76.18 $\rightarrow$ 75.48 & -0.70 & 2.66E9 (34.9\%) & 17.90E6 (30.1\%) \\
    & LeGR \citep{chin2019legr} & 76.10 $\rightarrow$ 75.70 & -0.40 & 2.37E9 (42.0\%) & - \\ 
    & \textbf{DI-SD}(Ours) & 76.10 $\rightarrow$ 76.10 & 0 & 2.31E9 (43.5\%) & 16.70E6 (34.8\%)\\
    & Hrank \citep{lin2020hrank} & 76.15 $\rightarrow$ 74.98 & -1.17 & 2.30E9 (43.8\%) & 16.15E6 (36.9\%) \\
    & NISP \citep{yu2018nisp} & - & -0.89 & 2.29E9 (44.0\%) & 14.38E6 (43.8\%) \\
    & Taylor-72 \citep{molchanov2019importance} & 76.18 $\rightarrow$ 74.50 & -1.68 & 2.25E9 (45.0\%) & 14.20E6 (44.5\%) \\
    & ThiNet-50 \citep{luo2017thinet} & 72.88 $\rightarrow$ 71.01 & -1.87 & 1.82E9 (55.8\%) & 12.40E6(51.6\%)\\
    & GDP \citep{lin2018accelerating} & 75.13 $\rightarrow$ 71.89 & -3.24 & 1.88E9 (54.0\%) & - \\
    & DCP \citep{zhuang2018discrimination} & 76.10 $\rightarrow$ 74.95 & -1.15 & 1.82E9(55.8\%) & 12.40M (51.6\%)\\
    & FPGM-40 \citep{he2019filter} & 76.15 $\rightarrow$ 74.83 & -1.32 & 1.90E9 (53.5\%) & - \\
    & \textbf{DI-unif}(Ours) & 76.10 $\rightarrow$ 75.42 & -0.68 & 1.80E9 (56.0\%) & 12.10E6 (52.7\%) \\
    \bottomrule
\end{tabular}
\label{tab:imagenet}
\vspace{-0.2in}
\end{table}
\endgroup

\section{Conclusion}
In this paper, we take a feature-map discriminant perspective for pruning deep neural networks. We develop theoretical guidelines to effectively quantify the feature-map discriminant power, giving rise to a novel Discriminant Information (DI) criterion. DI-based pruning removes channels with minimum influence to the DI value, which is measured as the derivative of DI w.r.t. the channel indicator variable. The versatility of DI also enables a intra-layer mixed precision quantization scheme to further compress the network. Moreover, we propose DI-based greedy pruning algorithm to deal with more realistic scenarios where we need to prune the network to meet certain resource budget. Our algorithm automatically decides the target pruned architecture in a highly efficient manner. To further accelerate our pruning process onnvery deep networks, we also propose a structure distillation technique to transfer the patterns of pruned structures from shallower networks to compressing deeper networks. Extensive experiments on various CNN architectures and benchmarks validates the effectiveness of our method.

\small
\bibliographystyle{abbrvnat}
\bibliography{ref}

\end{document}